\pdfoutput=1
\documentclass[11pt]{article}

\usepackage[final]{acl}

\usepackage{times}
\usepackage{latexsym}

\usepackage{tcolorbox}
\usepackage{soul}

\usepackage[T1]{fontenc}
\usepackage[utf8]{inputenc}

\usepackage{microtype}

\usepackage{inconsolata}

\usepackage{graphicx}
\usepackage{xcolor}
\usepackage{soul}
\usepackage{textcomp}

\definecolor{red-highlight}{RGB}{255, 153, 153}
\definecolor{blue-highlight}{RGB}{204, 229, 255}
\definecolor{yellow-highlight}{RGB}{255, 255, 204}
\definecolor{gray-highlight}{RGB}{224, 224, 224}
\definecolor{violet-highlight}{RGB}{153, 153, 224}
\sethlcolor{mycolor}

\title{\textcolor{red-highlight}{SAGE}val: The frontiers of \textcolor{red-highlight}{S}atisfactory \textcolor{red-highlight}{A}gent based NL\textcolor{red-highlight}{G} \textcolor{red-highlight}{E}valuation for reference-free open-ended text}

\author{
  \textbf{Reshmi Ghosh},
  \textbf{Tianyi Yao},
  \textbf{Lizzy Chen},
  \textbf{Sadid Hasan},
  \textbf{Tianwei Chen},
  \textbf{Dario Bernal},\\
  \textbf{Huitian Jiao},
  \textbf{H M Sajjad Hossain}
\\
  Microsoft
\\
  \small{
    \textbf{Correspondence:} \href{mailto:reshmighosh@microsoft.com}{reshmighosh@microsoft.com}
 }
}

\begin{document}
\maketitle
\begin{abstract}
Large Language Model (LLM) integrations into applications like Microsoft365 suite and Google Workspace for creating/processing documents, emails, presentations, etc. has led to considerable enhancements in productivity and time savings. But as these integrations become more more complex,  it is paramount to ensure that the quality of output from the LLM-integrated applications are relevant and appropriate for use. Identifying the need to develop robust evaluation approaches for natural language generation, wherein references/ground labels doesn't exist or isn't amply available, this paper introduces a novel framework called \verb|SAGEval| which utilizes a critiquing Agent to provide feedback on scores generated by LLM evaluators. We show that the critiquing Agent is able to rectify scores from LLM evaluators, in absence of references/ground-truth labels, thereby reducing the need for labeled data even for complex NLG evaluation scenarios, like the generation of JSON-structured forms/surveys with responses in different styles like multiple choice, likert ratings, single choice questions, etc.   %even for comp   the current state of natural language text evaluation for cases where LLM %Integrating Large Language Models (LLMs) in applications require comprehensive examination of intermediate as well as final responses to ensure quality. %Current benchmarks for evaluation are   can incur substantial costs,which has prompted recent advances in inference system optimizations. Today, these systems are evaluated against conventional latency and throughput metrics (eg. TTFT, TBT, Normalised Latency and TPOT). However, these metrics fail to fully capture the nuances of LLM inference, leading to an incomplete assessment of user-facing performance crucial for real-time applications such as chat and translation. In this paper, we first identify the pitfalls of current performance metrics in evaluating LLM inference systems. We then propose Metron, a comprehensive performance evaluation framework that includes fluidity-index– a novel metric designed to reflect the intricacies of the LLM inference process and its impact on real-time user experience. Finally, we evaluate various existing open-source platforms and model-as-a-service offerings using Metron, discussing their strengthsand weaknesses. Metron is availabl
\end{abstract}

\section{Introduction}
%\reshmi{Large Language Models (LLM) have opened up new avenues for natural language text generation and modification, which previously  }
Large Language Models (LLMs) have opened up new avenues for enhancing productivity \cite{Weise_Grant_2023b} and the scenarios where these models are utilized have gone from simple summarizing, translation, rewriting, Q/A tasks, to complex scenarios such as richly formatted text generation, code generation, involved creative writing tasks (such as open-ended story telling, intent specific list generation, open ended question generation for quizzes/surveys, etc,), and many more. LLM agents \cite{wu2023autogen}\cite{li2023camel} and application development are also becoming sophisticated by the day with the utilization of retrieval augmented generation techniques\cite{lewis2020retrieval}\cite{wadhwa2024rags}, where the response/output from one LLM, acts as intermediate input for another LLM, for downstream processing. The integration of black-box many proprietary LLMs \cite{achiam2023gpt},\cite{team2023gemini} in applications with advanced prompting methods and tooling\cite{liu2024toolnet}, is requiring the community to think about assessing quality \textbf{at all steps}, and not limiting to only analyzing the quality final output received from a Artificial Intelligence(AI) based product/application. Thus, an approach to analyze the quality of intermediate NLG texts produced by LLMs and applications are becoming important. \\

Interestingly, the past year witnessed a rise in the use of using LLMs to scale the evaluation of open-form and closed-form Natural Language Generation (NLG). NLG evaluation typically includes evaluating the generated text on multiple dimensions \cite{lin2023llm}, to obtain a comprehensive assessment about the inferred and generated content by auto-regressive models. %\reshmi{Modify later as this is copied until the next todo} However, multi-aspect evaluation remains challenging as it may require the evaluator to generalize to any given evaluation aspect even if it’s absent during training. In this paper, we introduce SAGE, a resource-efficient agent-based evaluation framework for open-form generations. %in both seen and un
%Single-agent frameworks with approaches like 
Self Consistency or Chain-of-Thought reasoning have been widely used for scaling evaluation work streams, and have shown promise, but are also emphasized on the need to fill the parity gap between human judgements and its current effectiveness. Parallely, researchers also questioned the use of LLM-based Evaluators\cite{panickssery2024llm}\cite{luong2024realistic} for analyzing the quality of natural language texts. But even woth the identified gaps, LLMs are the go-to approach, for a scalable automatic evaluation method for natural language text, that doesn't depend on human-annotators \cite{chen2024roleinteract, gao2024llm, saha2023branch, hada2023large}.\\

Simulatenously, agentic frameworks \cite{wu2023autogen} have enabled use of roles\cite{hong2023metagpt, li2023metaagents, li2023camel}, tools \cite{rasheed2024codepori, yang2024how2comm}, to solve complex tasks that have sub-optimal results only using a prompted call to an LLM. This popularity has enabled the exploration of agentic frameworks in LLM-based automatic evaluation of natural language text \cite{chan2023chateval, li2024mateval}, but these approaches still do not solve/examine scenarios of natural language text evaluation where there is no reference text/grounding data for the LLM evaluators to reference, while analyzing the text.   %\reshmi{reshmi to edit from here:\textsc{SAGEval} consists of of two steps: the first step involves the utilization of Chain-of-Thought reasoning for evaluating the quality of open-ended text on several pre-defined criteria and assigning a score. The second step: recognizing that the human evaluation process often involves seeking feedback and discussion with a collaborator (or a \emph{meta-reviewer}) known for their wisdom, involves calling a second agent, which we call \textsc{SAGE} agent, suggesting the addition of any metrics, revisiting the scores and finessing the first agent's explanations, thus improving the overall evaluation for NLG tasks.}\\
%istruction tuning stage that improves the model’s ability to follow evaluation instructions, and an enhanced instruction tuning stage that exploits the connections between fine-grained evaluation aspects to better assess text quality. To support the training of X-EVAL, we collect ASPECTINSTRUCT, the first instruction tuning dataset tailored for multi-aspect NLG evaluation spanning 27 diverse evaluation aspects with 65 tasks. To enhance task diversity, we devise an augmentation strategy that converts human rating annotations into diverse forms of NLG evaluation tasks, including scoring, comparison, ranking, and Boolean question answering. Extensive experiments across three essential categories of NLG tasks: dialogue generation, summarization, and data-to-text coupled with 21 aspects in meta-evaluation, demonstrate that our X-EVAL enables even a lightweight language model to achieve a comparable if not higher correlation with human judgments compared to the state-of-the-art NLGevaluators, such as GPT-4
\begin{figure}[t]
\centering
%\hspace{}
  \includegraphics[width=1.1\linewidth]{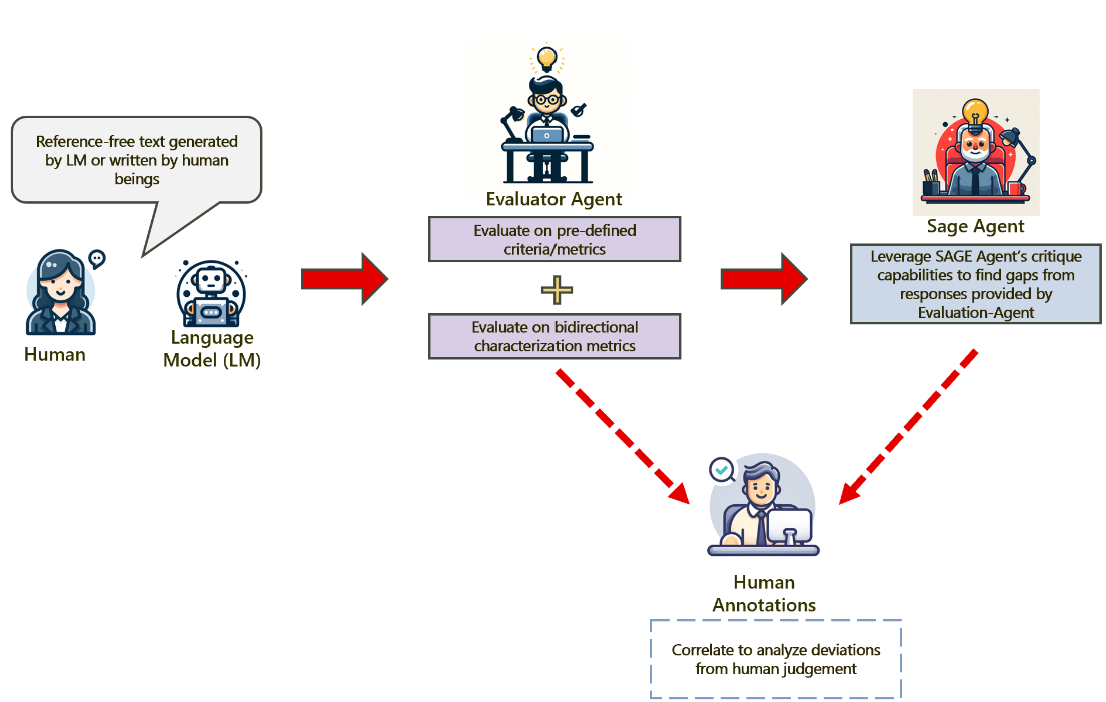} %\hfill
  \caption {\textbf{SAGEval} framework. SAGEval engages with a "wiser" role-based agent to validate scores assigned by the first LLM Evaluator for reference-free texts. }
\end{figure}
%Recently, the use of pre-trained language models for NLG evaluatio\textbf{}\\

In this paper, we push the boundaries of evaluation approaches leveraging LLMs to completely utilize their ability to judge and provide feedback, specifically for the case of applications, where generated text is open-ended and reference-free. We set up the \verb|SAGEval| framework to use in-context learning (few-shot) exemplars, self-reflecting on the judgements provided like established approaches, such as G-Eval\cite{liu2023gpteval} and GPT-score\cite{fu2023gptscore}, but also utilize a role-based agent for meta-evaluation, that is to provide feedback on assigned scores by reasoning and critiquing when needed. The meta-evaluator agent also suggests changes in the pre-defined scoring criteria to efficiently adapt to the open-ended nature of the generated text in absence of ground truth. This allowed to receive rectified/corrected scores from the critiquing agent, while also getting suggestions on how to improve the scoring criteria. From the \verb|SAGEval| framework, we also seek suggestions on new scoring criteria to compensate for gaps in pre-defined scoring criteria, which effectively allows us to 

In particular, this main contributions of our paper are:
\begin{itemize}
    \item We propose a new scalable framework of role-based LLM Agents for evaluating open-ended reference-free text that aligns better with human preferences when compared with other established approaches of leveraging LLMs as evaluators
    \item Through the proposed framework, we showcase the ability of LLM Evaluators to assume a role and critique scores and close gaps in scores generated by popular LLM Evaluator methods such as G-Eval, when reference-documents are not available for validation. Additionally, we also release the dataset and associated human annotations curated for ease of reproducibility.
    \item We demonstrate the capabilities of LLM Evaluators to not only score natural language text, but also propose new aspects for scoring comprehensively and increase coverage for evaluations.
\end{itemize}

\section{Related Works}
\label{related_work}
The popularity of Large Language Models (LLMs) has shifted the focus on the importance of understanding the quality of natural language generations and accelerating research for reference-free evaluation. Traditional metrics used for assessing the quality and correctness of natural language like BLEU (Bilingual Evaluation Understudy score; \cite{papineni2002bleu}), ROUGE (Recall-Oriented Understudy for Gisting Evaluation), and WER (Word Error Rate) \cite{klakow2002testing} to name a few, require well-calibrated and annotated ground-truth labels or refrences, which limits the scale of evaluation and slows the evaluation process. \\

LLMs have also exhibited impressive judging and evaluation capabilities \cite{qin2023chatgpt}\cite{wang2023chatgpt}\cite{chiang2023can}\cite{bubeck2023sparks} leading to the innovation of approaches such as: G-Eval \cite{liu2023gpteval}, first of it's kind unique framework for \textit{scaling} NLG evaluation reliably, which proved various reference-based metrics such as BLEU, ROUGE, etc., are insufficient in capturing contextual relevance and nuanced discrepancies unlike human annotators. G-Eval concluded that advanced models like GPT4 \cite{achiam2023gpt} were able to critique and identify gaps in NLG evaluation better, further closing the gap against human judgements. Since then many evaluation frameworks have been proposed and researchers have tried underscoring gaps in the known approaches. New approaches such as CheckEval\cite{lee2024checkeval}, LLM-Eval\cite{lin2023llm}, GPT-Score\cite{fu2023gptscore}, FreeEval\cite{yu2024freeeval}, and \textit{MMVet} \cite{yu2023mm}, MEGA\cite{ahuja2023mega}, Megaverse\cite{ahuja2023megaverse}, have tackled various evaluation tasks for various  natural language texts, but all these frameworks consider some form of context or incorporate reference documents to validate the quality of generated text.\\

%Simultaneously, the domain of evaluation has also 

Researchers have also elicited the nuanced gaps in proposed evaluation frameworks, discovering positional bias in LLM evaluators \cite{wang2023large}, and have tried augmenting LLM evaluation approaches by using multi-agent farmeworks\cite{wu2023autogen}\cite{chen2023agentverse}\cite{li2023camel}. This led to development of approaches such as Branch-Solve-Merge\cite{saha2023branch} and ChatEval\cite{chan2023chateval}, but these frameworks were catering tasks that had access to reference-documents.\\

We found that, there have been attempts to address NLG evaluation challenges for closed-book questions by introducing TrustScore \cite{zheng2024trustscore} and reference-free text in the form of dialogue-generation, story-generation, and paraphrase generation \cite{chen2023exploring}. A multi-agent framework, MATEval \cite{li2024mateval}, also was proposed to understand efficacy of evaluating \textit{Openmeva} \cite{guan-etal-2021-openmeva}. Although all these studies claimed to solve the challenge of evaluating NLG texts in absence of references, the tasks were primarily centered around evaluating text that is continuous and in the form of a paragraph or excerpt (in stories, dialogues, and paraphrases), and the nature of text that involves generation of lists, or a set of formatted questions, with response choices around a \textit{central theme}, is very different from the aforementioned tasks, as it lacks \textbf{direct continuity}.\\

Thus, in this paper, we introduce \verb|SAGEval|, in an attempt to address the gaps in open-ended reference-free NLG text types that is not continuous and centered around a "theme" and validate it's efficacy by demonstrating it's closeness to human annotations.

\section{Open-ended reference free text}
\begin{figure}[h!]
\centering
  \includegraphics[width=1.1\linewidth]{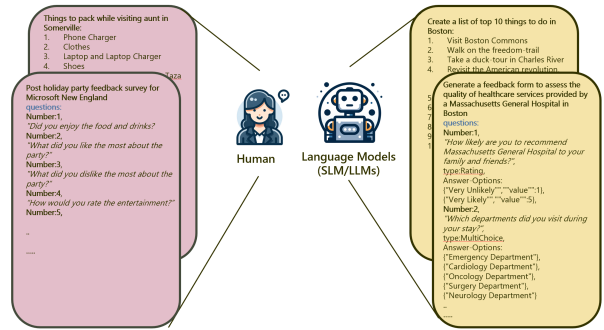} %\hfill
  \caption {\small{Open-ended human-drafted and NLG texts like lists, surveys, forms, contains sub-items or entities that are associated with a \textbf{central theme} such as "List of things to pack while traveling", or "Survey on assessing the quality of healthcare services", but these items (bullets in a list, questions in a survey) differ from each other, and it is important to make sure that the variance in open-ended text is coherent and aligned to the central theme.}}
  \label{fig:referencefreetext}
\end{figure}

To the best of our knowledge, there exists no open sourced reference-free open-ended NLG text, that are lists/surveys/forms with associated human annotations. Popular benchmarks used for assessing efficacy of new evaluation frameworks such as SummEval\cite{fabbri2021summeval}, QAGS\cite{wang2020asking}, Openmeva\cite{guan-etal-2021-openmeva}, GSM8K\cite{cobbe2021training}, MixedQA\cite{zheng2024trustscore}, etc. are not aligned with the problem that we are trying to solve.\\
Our inspiration to generate this dataset stems from %evaluation framework for reference-free text, we introduce a new benchmark and dataset for evaluating reference-free text questions generated as part of creating feedback forms/surveys/quizzes. \todo[inline, color=red!50]{add a statement starting we find that contemporary language models have an especially high tendency to hallucinate factually inaccurate biographies, often misrepresenting the relevant institutions and dattes}.
%comprehensive benchmarking framework to assess its capabilities 
 established products that support reference-free open-ended text generation like Google Forms, Microsoft Forms, Survey Monkey. Integration of Artificial Intelligence to these platforms (such as Microsoft Forms CoPilot \cite{Microsoft_2024}) to automate form/survey generation, requires robustly examining the outputs from the language models before surfacing it to the users.\\
 Thus, we introduce an unique benchmark for reference-free dataset containing 96 surveys and forms each centered around a different topic, that was generated by GPT-3.5-Turbo 0613 using a user prompt of at max 50 words. We also curated annotations from humans to score the generated surveys qualitatively across a pre-defined set of scoring criteria, as explained in section \ref{human-annotations}.
 
 Through rigorous benchmarking, we aim to provide a comprehensive assessment of \verb|SAGEval|'s capabilities and identify areas for further improvement in the pursuit of advancing natural language understanding and generation systems for \textbf{practical use}.

\section{SAGEval Framework}
We introduce \verb|SAGEval|, a new evaluation framework for open-ended reference-free text, that leverages the ability of role-based LLM Evaluator Agents to critique and expose gaps in scores assigned in absence of ground-truth references for comparison. The framework has two LLM agents, that objectively examines each instance of an open-ended reference free text, against a set of pre-defined aspects described in Section \ref{scoring-criteria}.

\subsection{Scoring Criteria for Aspects}\label{scoring-criteria}
Inspired from the \textit{aspects} leveraged in GPT-Score\cite{fu2023gptscore} and X-eval\cite{liu2023x}, we expand the criteria \textit{typically} used by LLM evaluators for scoring, and do not limit the Agents to judge on \textbf{Coherence}, \textbf{Fluency}, \textbf{Relevancy}, and \textbf{Consistency} only. This allows us to compensate for the lack of reference-data or ground-truth and perform a comprehensive evaluation of open-ended surveys and forms, while ensuring adherence to the intended/chosen theme of survey generation. \\

For every form/survey $x$ we pre-define
8 set of evaluation aspects $A$ (Accuracy, Semantic Diversity, Coherence, Relevancy, Audience Understandability, Audience Engagement, Fairness, Sentiment/Tone type). The description of each \textit{predefined aspect} is in Appendix \ref{Appdx:scoring}. The evaluation task is then formulated as:
$c = f (x, S, a)$, where $a \in A$ is the fine-grained aspect to be evaluated, and $f (\bullet)$ is the scoring function that provides an assessment $c$ w.r.t. the aspect $a$.

\subsection{Evaluator Agent}
The \verb|Evaluator Agent| is \textbf{based on the principles of G-Eval} \cite{liu2023gpteval}, which utilizes four templatizied sub-components: 1) a prompt that describes the evaluation task and expectation of the role of the \verb|Evaluator Agent|, 2) description of pre-defines aspects for assigning a score between 1-5 points) chain-of-thoughts (CoT) reasoning based guidance to execute the evaluation task (Appendix \ref{Appdx:scoring}), while also ensuring the \verb|Evaluator Agent| provides reasoning for assigning a particular score, and finally 3) exemplars for In-Context learning on how to format the response. %The prompt used for \verb|Evaluator Agent| is described in Appendix \ref{appendix-EvaluatorAgent} for easy reproducibility\\ 

This step in the SAGEval framework, is a \textit{first-pass} towards assigning scores for open-ended text, in absence of references.

\subsection{SAGE Agent}
After the \verb|Evaluator Agent| takes the first pass, the SAGEval framework utilizes a meta-evaluator agent called \verb|SAGE Agent|, assessing
the scores generated by the \verb|Evaluator Agent| and provides feedback. \verb|SAGE Agent| being the meta-evaluator is designed to objectively look at each instance of open-ended reference-free text ($x$) (here a form/survey) to:
\begin{enumerate}
    \item provide insights on how to rectify assigned scores by \verb|Evaluator Agent| on pre-defined aspects ($a \in A$)
    \item mutually exclusively provide suggestions to modify definitions of pre-defined aspects ($a$)
    \item and optionally suggest new aspects ($a$) to increase coverage of evaluation, and measure gaps, that pre-defined aspects fail to measure
\end{enumerate}
This setup is inspired from the humane need of seeking feedback from subject matter experts as part of a strategy to cross-examine scores, for example:, in any review process of scholarly articles, meta-reviewers
provide the finalized feedback input after reviewing feedback notes from individual reviewers. \\
And although established multi-agent framework\cite{chan2023chateval} often involves seeking feedback from multiple agents of different
competencies, we try to strike a balance between invoking multiple agents that gets hard to pro-
ductionize due to cost of utilizing many calls to LLMs, and iterating over feedback derived from
LLM-based evaluators.

\subsection{Preliminaries}

As proposed in G-Eval\cite{liu2023gpteval}, for both the evaluator Agents: \verb|Evaluator Agent| and \verb|SAGE Agent|, similar to \cite{liu2023gpteval}, we ensure that we tackle any skew of score distributions and ties in scores assigned for each pre-defined criteria, by normalizing the scores using probabilities of output tokens from the LLMs. Thus, for a given a set of scores by \verb|Evaluator Agent| and \verb|SAGE Agent| (from 1 to 5)
$S = {s_1, s_2, ..., s_n}$, the probability of each score $p(s_i)$ is calculated as: \begin{equation}
   score =  \sum_{i=1}^n p(s_i) \times s_i 
\end{equation}

\section{Human Annotations}\label{human-annotations}
\begin{figure*}[t]
%\centering
  \includegraphics[width=\linewidth]{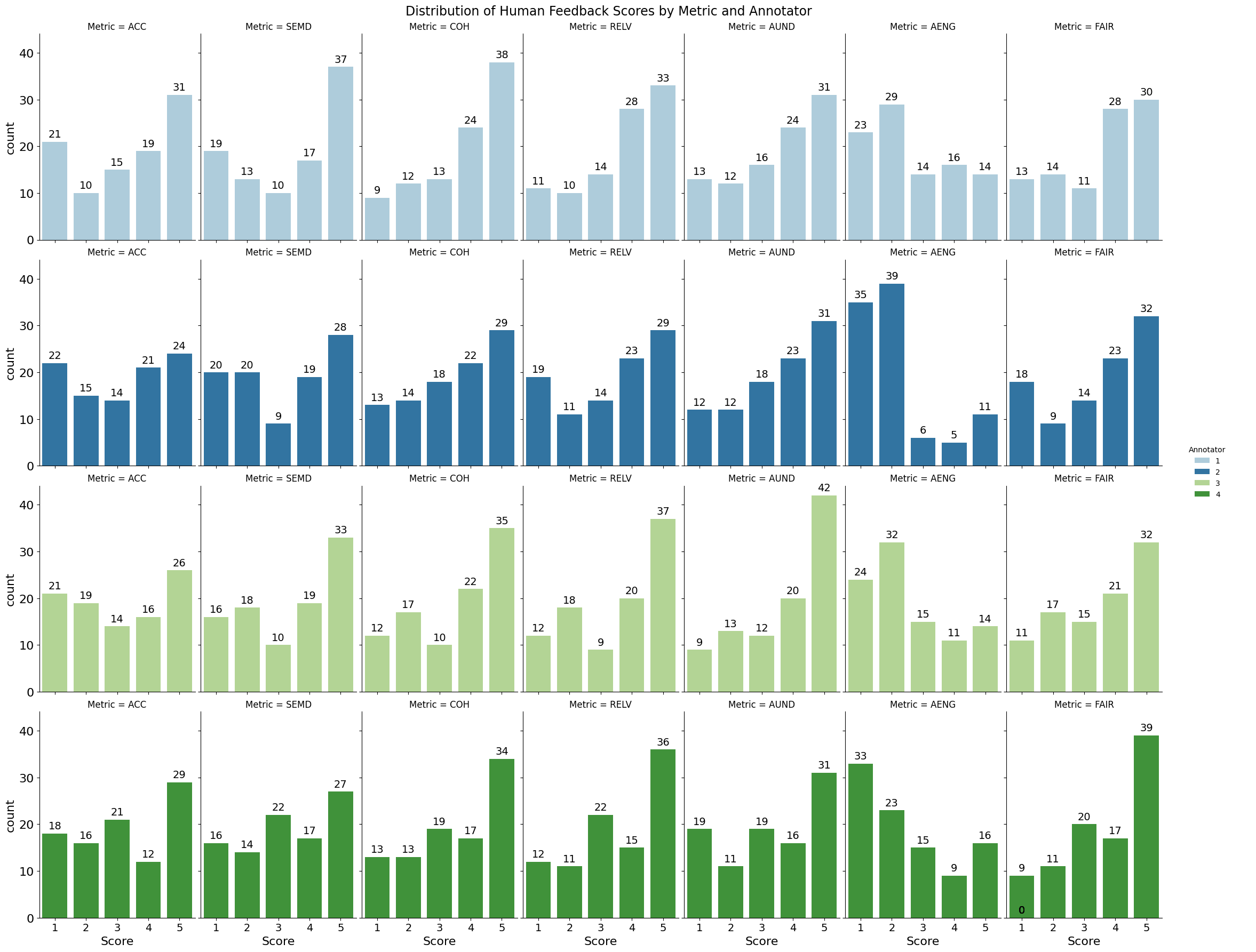} 
  \caption {Distribution of annotation scores (between 1-5) assigned to each Scoring Criteria: , by 4 highly experienced linguists who are experience with artificial intelligence. We note that, for the aspect \textit{Audience Engagement}, there is a dramatic shift in scores which heavily leans towards being low across (1 and 2) for all 4 human annotators. }\label{scoring_metrics}
\end{figure*}

The 96 open-ended surveys/quizzes/forms generated by GPT-3.5-Turbo 0613, were annotated by \textbf{4} highly experienced linguists who are familiar with Artificial Intelligence and were well-equipped to assess the quality of the responses. We collected these annotations for each scoring criteria defined in Appendix \ref{Appdx:scoring} and distribution of the annotations assigned (scores between 1-5) by the linguists are represented in Figure \ref{scoring_metrics}. Amongst all pre-defined scoring criteria we note that, evidently, the \textit{Audience Engagement} criteria had lower scores.

\begin{table}[ht!]
  \centering
  \begin{tabular}{lcccc}
    \hline
    \textbf{\footnotesize{Scoring Criteria}} & \textbf{\footnotesize{Neg}} & \textbf{\footnotesize{Pos}} & \textbf{\footnotesize{Total}} & \textbf{\footnotesize{Definition}} \\
    \hline
    \footnotesize{Accuracy}  & \footnotesize{41} & \footnotesize{0}  & \footnotesize{41}  & \footnotesize{\colorbox{blue-highlight}{2}} \\
    \hline
    \footnotesize{Semantic} \\ \footnotesize{Diversity} & \footnotesize{31} & \footnotesize{0}  & \footnotesize{31} & \footnotesize{12} \\
    \hline
    \footnotesize{Cohesion}  & \footnotesize{15} & \footnotesize{0}  & \footnotesize{15}  & \footnotesize{\colorbox{blue-highlight}{2}} \\
    \hline
    \footnotesize{Relevancy} & \footnotesize{15} & \footnotesize{0}  & \footnotesize{15}  & \footnotesize{4} \\
    \hline
    \footnotesize{Audience}\\ \footnotesize{Understandability} & \footnotesize{9}  & \footnotesize{1}  & \footnotesize{10} & \footnotesize{11} \\
    \hline
    \footnotesize{Audience}\\ \footnotesize{Engagement} & \footnotesize{76} & \footnotesize{5}  & \footnotesize{\colorbox{red-highlight}{81}} & \footnotesize{\colorbox{red-highlight}{11}} \\
    \hline
    \footnotesize{Fairness} & \footnotesize{5}  & \footnotesize{1}  & \footnotesize{\colorbox{blue-highlight}{6}}  & \footnotesize{7} \\
    \hline
    \footnotesize{Sentiment} & \footnotesize{1}  & \footnotesize{10} & \footnotesize{11} & \footnotesize{\colorbox{red-highlight}{32}} \\
    \hline
  \end{tabular}
  \caption{Number of rectifications in scores as suggested by the \textbf{SAGE Agent} broken down as Neg(Negative; decrease in scores $\big\downarrow$), Pos(Positive; increase in scores $\big\uparrow$), Total (Negative + Positive). \textbf{Defintion} column indicates, number of instances across the proposed 96 reference-free form/surveys dataset, \textbf{SAGE Agent} proposed changes in the way the scoring criteria was defined per criteria. We note that, \textit{Audience Engagement} had the largest number of score disagreements between SAGE Agent and Evaluator Agent, with \textit{Fairness} being the minimum. For rectifications in scoring criteria definitions, \textit{Sentiment} criteria had the largest number of disagreements. Blue highlights represents the lower range of disagreements and red highlights represents the upper most range of disagreements made by \textsc{SAGE Agent} across all aspects.}
  \label{tab:disagreement_metrics}
\end{table}

\section{Experiments and Results}

To assess the effectiveness of our SAGEval framework, we utilize the new introduced benchmark of 96 open-ended surveys/forms/quizzes and evaluate the framework against existing reference-free methods of evaluation, i.e, popular LLM evaluation frameworks such as \textbf{G-Eval}, \textbf{CheckEval}, \textbf{FreeEval}, and \textbf{MATEval}. Additionally, for \verb|SAGE Eval|, we test out three versions, vanilla SAGEval, Self-Reflection (\textbf{SR}), and Chain-of-Thought (\textbf{CoT}) incorporated \verb|SAGEval|.\\

%And as stated earlier, the \verb|Evaluator Agent|, within \verb|SAGEval| framework, is an adaptation of \textbf{G-Eval}, and thus in Table \ref{Tab1:results}, we report out G-Eval scores in the row of LLM-Eval.  %\reshmi{three of the authors manually annotate the “win/tie/lose” outcomes of responses from ChatGPT and Vicuna-13B independently in all 80 Vicuna Benchmark questions. All of the annotators Following the same template as the original Vicuna, the annotators are instructed to assess the responses provided by Vicuna-13B and ChatGPT from four different perspectives: helpfulness, relevance, accuracy, and level of detail. The responses of Vicuna and ChatGPT are presented to the annotators in random order. The evaluation process for each example took an average of three minutes. The final result is based on the majority opinion among three annotators.}
%\section{Evaluating reference-free text using LLM Evaluator}
%We report results on reference-free text evaluation and discuss the major findings about LLM evaluators' efficiency to critique text without references or grounding data.

We do not utilize any scoring mechanism dependent on ground-truth, for examining the effectiveness of SAGEval framework. This is because metrics like BLEU, ROUGE, METEOR, etc., are dependent on comparison against reference documents, which doesn't align with the goal of this body of work. 

\begin{figure*}[t!]
\centering
\hspace{-10mm}
  \includegraphics[width=1.05\linewidth]{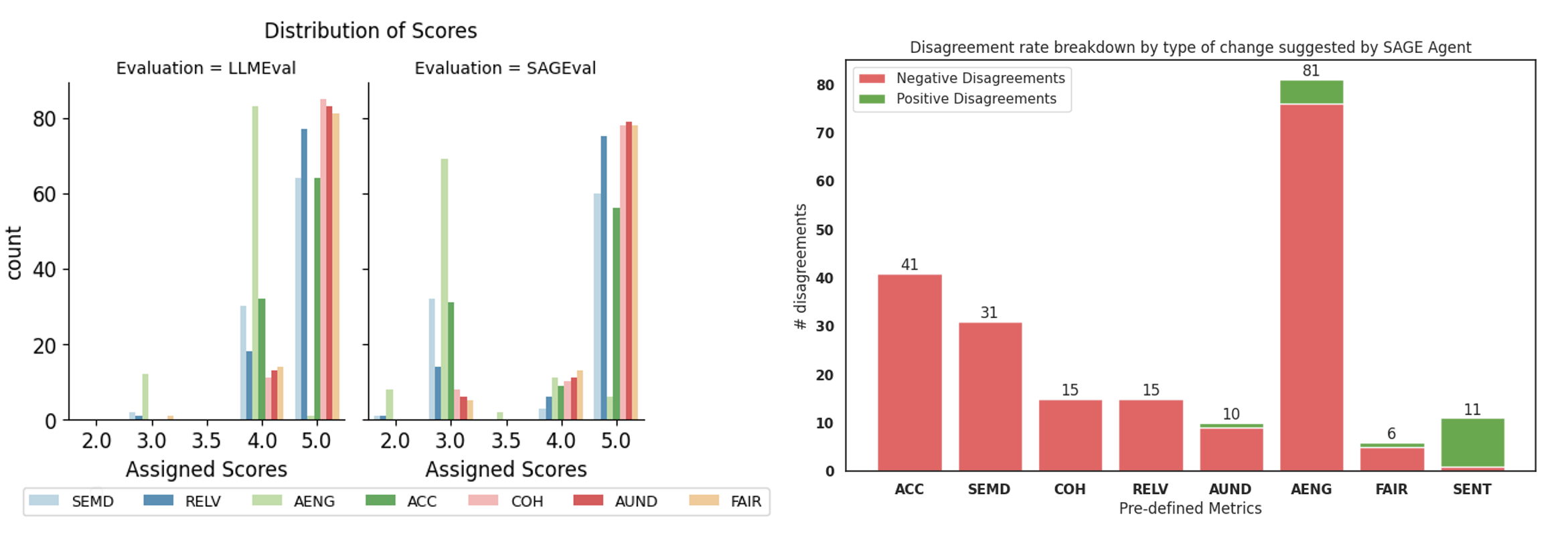} %\hfill
  \caption {\small{Scores distribution by \textit{SAGE Agent} compared scores assigned by \textit{Evaluator Agent}. We find that \textbf{Evaluator Agent} is inclined towards assigning higher ratings (4s and 5s) across all criteria, whereas \textbf{SAGE Agent} is more critical and pushes the score distribution towards 3s and a couple of 2s. }}
  \label{fig:distribution_of_scores}
\end{figure*}

\subsection{Finding 1: A role-based agent to critique, rectifies LLM Evaluator scores}
With the introduction of critiquing \verb|SAGE Agent| for rectifying scores assigned by \verb|Evaluator Agent| in absence of any references for open-ended forms and surveys that were generated by another LLM, the distribution of scores across all scoring criteria changes. Figure \ref{fig:distribution_of_scores} (a) compares the distribution of assigned scores between \verb|Evaluator Agent| (based on G-Eval) and \verb|SAGE Agent|, and we find that the distribution of scores become less heavier on 4s and shift towards 2s and 3s. Additionally, in Figure \ref{fig:distribution_of_scores} (b) we quantify the number of times per scoring criteria \verb|SAGE Agent| changes the magnitude and direction of scores assigned by \verb|Evaluator Agent|. We define \textbf{direction}, as either increasing or decreasing from original score. We find that, for $\sim$92\% of total score rectifications made, \verb|SAGE Agent| \textbf{negatively disagreed} with \verb|Evaluator Agent|, and the corrected scores were smaller than original scores, that's why the shift in score values from 5s and 4s $\longrightarrow$ 3s and 2s.\\
Interestingly, \textbf{Audience Engagement} scoring criteria had the most number of disagreements, wherein \verb|SAGE Agent| suggested 81 score changes in total, with 76/96 times to lower the assigned scores by \verb|Evaluator Agent|, and 5/96 times increased the scores to be increased (positive disagreement; $\big\uparrow$). This was followed by \textbf{Accuracy}, \verb|SAGE Agent| rectified 41/96 instances by lowering the scores, i.e., negatively disagreeing with \verb|Evaluator Agent|.\\
In addition, to the rectification of scores, \verb|SAGE Agent| also suggests correction to the aspect \textbf{definitions}, if it believes that the pre-defined aspect definitions do not comprehensively cover the scoring criteria defined. We find that across the 96 data points, \textbf{\textit{Sentiment/Type}} aspect identified by \verb|SAGE Agent| in the \textsc{SAGEval} framework to not approrpiately examine the surveys/forms for 32 times.

%% trial
\begin{table*}
    \centering
    \begin{tabular}{p {1.5cm} |c  c | c c | c c | c c| c c | c c | c c }
%&
%\multicolumn{2}{c|}{\textbf{Cross Segment BERT}} 
%&
%\multicolumn{2}{c|}{\textbf{Hierarchical Bi-LSTM}}
%&
    \footnotesize{\textbf{Scoring Criteria}} &
    \multicolumn{2}{p{0.8cm}|}{\footnotesize{\textbf{ACC}}} &
    \multicolumn{2}{p{0.4cm}|}{ \footnotesize{\textbf{SEMD}}} &
    \multicolumn{2}{p{0.4cm}|}{ \footnotesize{\textbf{COH}}} &
    \multicolumn{2}{p{0.4cm}|}{ \footnotesize{\textbf{RELEV}}} &
    \multicolumn{2}{p{0.4cm}|}{ \footnotesize{\textbf{AUND}}} &
    \multicolumn{2}{p{0.4cm}|}{ \footnotesize{\textbf{AENG}}} &
    \multicolumn{2}{p{0.4cm}}{ \footnotesize{\textbf{FAIR}}} \\
    &
    \footnotesize \textbf{$\rho$} &
    \footnotesize \textbf{$\tau$} &
      \footnotesize \textbf{$\rho$} &
    \footnotesize \textbf{$\tau$} &
      \footnotesize \textbf{$\rho$} &
    \footnotesize \textbf{$\tau$} &
      \footnotesize \textbf{$\rho$} &
    \footnotesize \textbf{$\tau$} &
      \footnotesize \textbf{$\rho$} &
    \footnotesize \textbf{$\tau$} &
      \footnotesize \textbf{$\rho$} &
    \footnotesize \textbf{$\tau$} &
      \footnotesize \textbf{$\rho$} &
    \footnotesize \textbf{$\tau$} 
\\
    %\footnotesize{\textbf{Scoring Criteria}} & \footnotesize{\textbf{Accuracy (ACC)}} & {} & \footnotesize{\textbf{Semantic Diversity (SEMD)}} & & \footnotesize{\textbf{Cohesion Score (COH)}} & & \footnotesize{\textbf{Relevancy Score (RELEV)}} & & \footnotesize{\textbf{Audience Understandability (AUND)}} & &\footnotesize{\textbf{Audience Engagement (AENG)}} & & \footnotesize{\textbf{Fairness (FAIR)}} & \\ 
%\footnotesize{Scoring Criteria & Accuracy & Semantic Diversity & Cohesion Score & Relevancy Score & Audience Understandability & Audience Engagement & Fairness} \\ 
    \hline
    \hline

    %\vspace{0.01mm}\\
    \footnotesize{G-Eval} & \scriptsize{0.49} & \scriptsize{0.44} &   \scriptsize{0.62} & \scriptsize{0.49} &  \scriptsize{0.32} & \scriptsize{0.43} &  \scriptsize{0.47} & \scriptsize{0.43} &   \scriptsize{0.33} & \scriptsize{0.40} & \scriptsize{0.25} & \scriptsize{0.35} &  \scriptsize{0.41} & \scriptsize{0.36}
    %\vspace{0.01mm}\\
    %\vspace{0.01mm}\\
    \\
    \footnotesize{CheckEval} & \scriptsize{0.43} & \scriptsize{0.42} &   \scriptsize{0.59} & \scriptsize{0.57} &  \scriptsize{0.28} & \scriptsize{0.27} &  \colorbox{red-highlight}{\scriptsize{0.40}} & \colorbox{red-highlight}{\scriptsize{0.39}} &   \scriptsize{0.29} & \scriptsize{0.28} & \scriptsize{0.21} & \scriptsize{0.23} &  \scriptsize{0.39} & \footnotesize{0.38}\\
    %\vspace{0.01mm}\\
    \footnotesize{ChatGPT-4o} & \scriptsize{0.37} & \scriptsize{0.33} &   \scriptsize{0.43} & \scriptsize{0.39} &  \scriptsize{0.26} & \scriptsize{0.27} &  \scriptsize{0.43} & \scriptsize{0.39} &   \scriptsize{0.36} & \scriptsize{0.34} & \scriptsize{0.21} & \scriptsize{0.25} &  \scriptsize{0.32} & \scriptsize{0.31}\\
    %\vspace{0.01mm}\\
    \footnotesize{FreeEval} & \colorbox{red-highlight}{\scriptsize{0.35}} & \colorbox{red-highlight}{\scriptsize{0.31}} &   \colorbox{red-highlight}{\scriptsize{0.38}} & \colorbox{red-highlight}{\scriptsize{0.38}} &  \scriptsize{0.24} & \scriptsize{0.23} &  \scriptsize{0.42} & \scriptsize{0.35} &   \colorbox{red-highlight}{\scriptsize{0.22}} & \colorbox{red-highlight}{\scriptsize{0.21}} & \colorbox{red-highlight}{\scriptsize{0.19}} & \colorbox{red-highlight}{\scriptsize{0.17}} &  \colorbox{red-highlight}{\scriptsize{0.31}} & \colorbox{red-highlight}{\scriptsize{0.30}}\\
    %\vspace{0.01mm}\\
    \footnotesize{MATEval} & \scriptsize{0.36} & \scriptsize{0.30} &   \scriptsize{0.42} & \scriptsize{0.39} &  \colorbox{red-highlight}{\scriptsize{0.22}} & \colorbox{red-highlight}{\scriptsize{0.21}} &  \scriptsize{0.41} & \scriptsize{0.43} &   \scriptsize{0.35} & \scriptsize{0.32} & \scriptsize{0.38} & \scriptsize{0.37} &  \scriptsize{0.40} & \scriptsize{0.35}\\
    %\vspace{0.01mm}\\
    %\footnotesize{MATEval} & \footnotesize{rho1} & \footnotesize{tau1} &   \footnotesize{rho1} & \footnotesize{tau1} &  \footnotesize{rho1} & \footnotesize{tau1} &  \footnotesize{rho1} & \footnotesize{tau1} &   \footnotesize{rho1} & \footnotesize{tau1} & \footnotesize{rho1} & \footnotesize{tau1} &  \footnotesize{rho1} & \footnotesize{tau1}\\
    %\vspace{0.01mm}\\
    \hline
    %\vspace{0.01mm}\\
    \footnotesize{\textbf{SAGEval}} & \colorbox{blue-highlight}{\scriptsize{0.63}} & \colorbox{blue-highlight}{\scriptsize{0.56}} &   \colorbox{blue-highlight}{\scriptsize{0.65}} & \colorbox{blue-highlight}{\scriptsize{0.57}} &  \colorbox{blue-highlight}{\scriptsize{0.41}} & \colorbox{blue-highlight}{\scriptsize{0.50}} &  \colorbox{blue-highlight}{\scriptsize{0.48}} & \colorbox{blue-highlight}{\scriptsize{0.48}} &   \colorbox{blue-highlight}{\scriptsize{0.44}} & \colorbox{blue-highlight}{\scriptsize{0.46}} & \colorbox{blue-highlight}{\scriptsize{0.49}} & \colorbox{blue-highlight}{\scriptsize{0.46}} &  \colorbox{blue-highlight}{\scriptsize{0.44}} & \colorbox{blue-highlight}{\scriptsize{0.45}}\\
    %\vspace{0.01mm}\\
    %\footnotesize{SR} & \footnotesize{0.62} & \footnotesize{0.55} &   \footnotesize{0.61} & \footnotesize{0.56} &  \footnotesize{0.39} & \footnotesize{0.47} &  \footnotesize{} & \footnotesize{tau1} &   \footnotesize{rho1} & \footnotesize{tau1} & \footnotesize{rho1} & \footnotesize{tau1} &  \footnotesize{rho1} & \footnotesize{tau1}\\
    %\vspace{0.01mm}\\
    %\footnotesize{CoT} & \footnotesize{rho1} & \footnotesize{tau1} &   \footnotesize{rho1} & \footnotesize{tau1} &  \footnotesize{rho1} & \footnotesize{tau1} &  \footnotesize{rho1} & \footnotesize{tau1} &   \footnotesize{rho1} & \footnotesize{tau1} & \footnotesize{rho1} & \footnotesize{tau1} &  \footnotesize{rho1} & \footnotesize{tau1}\\
    \hline
    
    \end{tabular}
    
    \caption{\small{Spearman ($\rho$) and Kendall-Tau ($\tau$) correlations of defined metrics on reference-free dataset. SAGEval outperforms LLMEval (which is based on G-EVal framework) on all aspects, \textbf{ACC:} Accuracy, \textbf{SEMD:}Semantic Diversity, \textbf{COH:}Coherence, \textbf{RELEV:}Relevancy, \textbf{AUND:} Audience Understandability, \textbf{AENG:} Audience Engagement, \textbf{FAIR:}Fairness. We observe that SAGEval has the largest values of correlations against human feedback, thus outperforming all other LLM evaluation techniques and NLP metrics. We also highlight the largest differences in correlation values across all criteria, with red being lowest and blue being the largest.}}
    \label{Tab1:results}
    
\end{table*}

%\subsection{Finding 2: In absence of reference grounding data, a critiquing Agent over an LLM evaluator, aligns better with human-judgements}
\subsection{Finding 2: Rectified scores from a critiquing Agent over aligns better with human-judgements}

We also conducted meta-correlation analysis over scores generated by popular evaluation frameworks leveraging LLMs, against the annotations by 4 linguists using \textit{Spearman Rank} ($\rho$) and \textit{Kendall Tau} ($\tau$) correlation. Table \ref{Tab1:results} validates the effectiveness of proposed \verb|SAGEval| framework, in comparison to CheckEval, ChatGPT-4o, FreeEval, MATEval, and G-Eval. Across all scoring criteria, utilizing \verb|SAGE Agent| to critique and correct scores assigned by LLM-Eval for open-ended reference-free text, \textbf{results in improved alignment with human annotators}. Particularly for Accuracy (ACC), Audience Understanding (AUND) and Audience Engagement (AENG), SAGEval framework results in correlation scores that are $\sim$20\%  more than LLM-Eval (G-Eval).\\
Across all proposed evaluation approaches using LLM evaluators and multiagent framework (MATEval) for reference-free text, \verb|SAGEval| framework achieves significantly better performance, as was clearly demonstrated by highest correlation with human feedback.

%\subsection{Finding 2:Human Agreement with SAGEval scores lies wit}

\begin{figure}[h!]
\centering
%\hspace{cm}
  \includegraphics[width=1.05\linewidth]{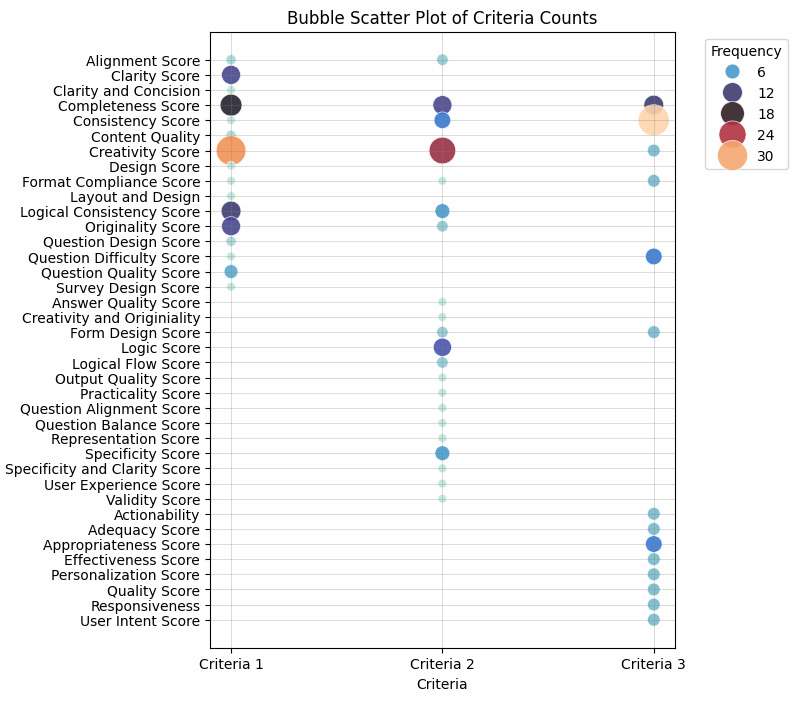} \hfill
  \caption {Term-topic frequency distributions of suggested aspects or scoring criteria (upto 3) by \textbf{SAGE Agent} for increasing evaluation coverage across 96 data points. We find that along with the pre-defined aspects, \textbf{SAGE Agents} suggests inclusion of Creativity Score and Content Quality Score for >40\% of all suggestions.}\label{fig:TopicModel}
\end{figure}

\subsection{Finding 3: Scoring Criteria gaps and alignment}
We recognize that the nature of open-ended reference-free text constitutes of sub-entities that may be different from each other. For example, in a feedback survey, each question of the survey will be associated with a \textbf{central theme} (such as customer service feedback form), but the questions itself will be different from each other, each separately trying to assess various aspect of "customer services". In such cases of open-ended reference-free natural language text, where there exists variance in generated text surrounding the  pre-defined scoring criteria may not always comprehensively judge the generated text, and there is a need for adding new \textit{aspects} for scoring and assessing better.\\

\verb|SAGE Agent| is prompted to also ensure if the pre-defined scoring criteria is For the scope of this paper, we assess comprehensively the efficacy of existing LLM evaluation frameworks/approaches on data and For the first time, we design an LLM Evaluator Agent to not only critique the natural language reference-free content on pre-defined scoring criteria, but also prompt to generate suggestions on additional aspects for scoring.\\
We find that for all 96 data points, \verb|SAGE Agent| suggests adding additional aspects for increasing evaluation coverage. We perform topic modeling across these suggestions to extract new aspects suggested as shown in Figure \ref{fig:TopicModel}. We find that \verb|SAGE Agent| across all 96 datapoints, repeatedly suggested inclusion of \textsc{Creativity Score} and \textsc{Content Quality Score} as the first, second, or third aspect suggestion to increase evaluation coverage.\\

This supplementary finding underscores the value of recognizing gaps in pre-defined aspects incorporate by developers/researchers for evaluation. This insight also demonstrates the value of not only using LLM evaluators for scoring aspects, but also showcases the capability of LLM Evaluator to assess gaps and suggest addition of new aspects \textbf{customized to the evaluation task}.

\section{Conclusion}
This is the first paper to comprehensively study open-ended reference free text and propose a framework comprising of a critiquing Agent to rectify scores and align more closely with human evaluations. We  confirm propose a new framework of evaluation that can comprehensively evaluate open-ended reference-free text generated by LLMs without labels. Evaluation approaches with minimal dependency on labels or reference text, opens up new avenue for LLM integration into products and also  
\\

\section{Limitations}
In this paper, we introduce \verb|SAGEval| framework, and comprehensively demonstrate the efficacy of this framework across various forms/surveys that were automatically generated by GPT-3.5-Turbo 0613, and the annotated by 4 experienced linguists. We focused on validating the efficacy of the framework on individual datapoint, that has a certain degree of variability, while also being structurally formatted. Although this can be considered as limited scope, and thus for future work, the efficacy of SAGEval on mores structured (JSON/Tables) and unstructured data (conversations/paragraphs of unrelated text generated via LLMs;) formats should be examined. We believe that the critiquing agent, together with the ability to evaluate on newer dimensions of scoring criteria in \verb|SAGEval|, can expose gaps in these datasets, like a human would, thereby increasing the usability of LLM-evaluators.

\section*{Acknowledgments}This work was completed with the help of Microsoft Forms team and Microsoft Office AI team. We are very thankful for all the support the liguists involved have provided on annotations, and the Microsoft Forms team on their requirements for structured data formats.
\bibliography{acl_latex}
\section{Appendices}
\appendix
\section{Scoring Criteria}\label{Appdx:scoring}
\textbf{Accuracy:} is defined analyzing output text and then tries to judge whether there are any inaccuracies, missing, or unfactual content with respect to the \verb|user prompt|, i.e., the original prompt intention.
The criteria suggests to:\\
    1. Read the generated output \verb|form/survey/quiz| text carefully and identify the main theme across all sections and questions, and option choices (for the case of multichoice and single choice questions).\\
    2. Check if the general theme of the content in form/survey/quiz is aligned to the theme of the prompt (user ask), and if it presents them in a clear and logical order.\\ 
    3. Assign a score for Accuracy on a scale of 1 to 5, where 1 is the lowest and 5 is the highest based on the Evaluation Criteria.\\
    \\
\textbf{Semantic Diversity:} This criteria looks at the generated output text and then tries to judge whether the questions across all sections (if present) and the form are diverse, meaning they are semantically different and there are no duplicates.
    Evaluation Steps for second Criteria: \\
    1. Read the generated output form/survey/quiz text carefully and ensure that there are no duplicates.\\
    2. Also check if the content in form/survey/quiz is semantically rich and aligns to the theme of the prompt (user ask), while being diverse/different from each other.\\
    3. Assign a score for Semantic Diversity on a scale of 1 to 5, where 1 is the lowest and 5 is the highest based on the Evaluation Criteria.\\
    \\
\textbf{Cohension:}
    This criteria looks at the generated output text and then tries to judge whether the questions across all sections (if present) and the form are fluent and are grammatically correct, meaning the title, description, questions, options (in case of single choice, multichoice and rating), section titles, and section description have no typos, or grammatical errors.
    Evaluation Steps for third Criteria: \\
    1. Read the generated output form/survey/quiz text carefully and ensure that there are no typos or grammatical errors.
    2. Also check if the content in form/survey/quiz is fluent in english and coherent to understand.\\
    3. Assign a score for Cohesion on a scale of 1 to 5, where 1 is the lowest and 5 is the highest based on the Evaluation Criteria.\\
    \\
\textbf{Relevancy:}
    Evaluation on fourth Criteria: 
    This criteria looks at the generated output text and then tries to judge whether the questions across all sections (if present) and the form are relevant with respect to the prompt (user ask)?\\
    Evaluation Steps for fourth Criteria: \\\
    1. Read the generated output form/survey/quiz text carefully and ensure that all questions, section titles and options are relevant and important to the "user ask".\\
    2. Assign a score for Relevancy on a scale of 1 to 5, where 1 is the lowest and 5 is the highest based on the Evaluation Criteria.\\
 \\
\textbf{Audience Understandability:}
    This criteria looks at the generated output text and then tries to judge whether the questions across all sections (if present) and the form would be understandable by the audience responding to the survey/quiz wwithout any further clarifications?\\
    Evaluation Steps for fifth Criteria:\\
    1. Assume that you (GPT4 model) are the responder of the form/survey/quiz generated, and now read the generated output form/survey/quiz text carefully.\\
    2. After reading through the contents of the form/survey/quiz generated, please assign a "Audience Understandability" score on a scale of 1 to 5, where 1 is the lowest and 5 is the highest based on the Evaluation Criteria.\\
    \\
    \textbf{Audience Engagement score:} 
    This criteria looks at the generated output text and then tries to judge whether the questions across all sections (if present) and the form would be engaging for the audience responding to the survey/quiz.
    Evaluation Steps for sixth Criteria: \\
    1. Assume that you (GPT4 model) are the responder of the form/survey/quiz generated, and now read the generated output form/survey/quiz text carefully.\\
    2. After reading through the contents of the form/survey/quiz generated, please assign a "Audience Engagement" score on a scale of 1 to 5, where 1 is the lowest and 5 is the highest based on the Evaluation Criteria.\\
    \\
    \textbf{Fairness score:} 
    This criteria looks at the generated output text and then tries to judge whether the questions across all sections (if present) and the form are fair and without any bias that may cause any form of discomfort to any section of the society, especially minority groups.\\
    Evaluation Steps for seventh Criteria: \\
    1. Read the generated output form/survey/quiz text carefully and ensure that all questions, section titles, title of the form, description of the form are generated in a language that is fair, without any bias, or harmful content, that may cause discomfort to the responders.\\
    2. Also check if the conetnt in form/survey/quiz should be flagged on any Responsible AI standards.\\
    3. Assign a score for Fairness on a scale of 1 to 5, where 1 is the lowest and 5 is the highest based on the Evaluation Criteria.\\
    \\
   \textbf{Sentiment/Tone type:} 
    This criteria looks at the generated output text and then tries to identify the sentiment of the content by analyzing the questions across all sections (if present) and the form.
    Evaluation Steps for eight Criteria: \\
    1. Read the generated output form/survey/quiz text carefully and identify from the language of all questions, section titles, title of the form, description of the form the sentiment it conveys.\\
    2. Unlike the previous evaluation criteria which were assign a score for Fairness on a scale of 1 to 5, here please output tone/sentiment of the generated content (questions).

"""
%\end{tcolorbox}

%\subsection{NLG by LLM:}\label{appendix-NLGgenerator}

\end{document}